\documentclass[11pt]{article}
\usepackage{cite}
\begin{document}
\author{Eugene Charniak\footnote{Thanks to George Kanidars and Michael Littman for  advice and guidance.  Problems, of course, are completely my own.}\\ Department of Computer Science \\ Brown University}
\title{Extrapolation in Gridworld Markov-Decision Processes}
\maketitle
\begin{abstract}
Extrapolation in reinforcement learning is the ability to generalize at test time given states  that could never have occurred at training time.  Here we consider
four factors that lead to improved extrapolation in a simple Gridworld environment: (a) avoiding maximum Q-value (or other deterministic methods) for action choice at test time, (b)
ego-centric representation of the Gridworld, (c) building rotational  and mirror symmetry into the learning mechanism using rotational and mirror invariant
convolution (rather than standard translation-invariant convolution), and (d) adding a maximum entropy term to the loss function to encourage
equally good actions to be chosen equally often.
\end{abstract}

\section{Introduction}

Reinforcement learning commonly is studied with no distinction made between training and testing data.   This is reasonable when the set of states encountered in training is so large that  the agent is unlikely to see a novel (test-time) situation thereafter or
if the research goals lie elsewhere, such as efficient policy convergence at
training time.  

However, for RL to be trusted in the ``wild,'' it is so important
that it works in novel cases that increasingly researchers deliberately distinguish between training
and testing data, and ensure that the latter includes, ``unreachable'' states\cite{witty2018measuring},
They are unreachable in the sense that they cannot be reached during training.   In the literature, generalization to unreachable states is
often called ``extrapolation'', as opposed to interpolation \cite{packer2018assessing}).
So, for example,
\cite{witty2018measuring}, 
test the extrapolation ability of RL polices on the Atari game Amidar, by starting test examples
with, e.g.,
one fewer ``enemy'' than those in training.  Since 
the MDP actions never add or remove enemies, such a state can never occur in training and thus are
unreachable.  The researchers then
show that the policies learned during training are disastrously bad on these states, such as causing the agent never to leave
the start position as it continuously  tries to move into a neighboring wall.  The present paper also
focuses on  of extrapolation --- handling unreachable
states at test time --- but we have chosen a much simpler environment, Gridword, as we believe its simplicity will make the challenges of extrapolation
easier to diagnose.

After specifying in more detail our class of Gridword MDPs (Section \ref{prelim}) we then explore methods to
improve unreachable state generalization:  first we explore policy-gradient methods vs. those based
on Q values (Section \ref{dqnrei}), then ego-centric environment representation (Section \ref{egosec}), 
exploiting symmetry (Section \ref{symsec}),
and adding a maximum entropy bias to action choices (Section \ref{entsec}). 

\section{Previous Work}

While a major impetus to this research was the work of \cite{witty2018measuring}, where the authors show how what appear to be simplifying
changes to a game at test time can cause the learned policy to profoundly malfunction, the previous work most similar to the current paper is that by \cite{zhang2018study}, where the authors also 
use Gridworld as the environment of choice.  They characterize extrapolation as the other extreme from overfitting.  From this point of view
they are more concerned with model capacity than we and keep to a single basic a3c\cite{williams1992simple} architecture, significantly varying the hyperparameters.   They also explored the effects of move randomization during training, and found that while it helped in extrapolation the effect was
not large.  The two modifications we find most critical in our discussion, ego-centric representation and rotational symmetry, are not considered in
that study.

Taking a slightly different tact, several recent works set out to create game environments particularly well suited for extrapolation studies.  \cite{nichol2018gotta}
have developed a version of the Sonic the Hedgehog\texttrademark  that allows the user to get at enough of the internals to create an environment in which different levels of the game are available either for training, or testing. Inspired by this new version of Sonic, \cite{cobbe2018quantifying} create a new game from scratch, {\em coin run}, designed to make training and testing on distinct levels easy.  Similarly \cite{juliani2019obstacle} create the
game {\em obstacle tower}.

\section{Preliminaries\label{prelim}}

In reinforcement learning, a Gridworld is a Markov-Decision Process 
in which a single agent moves around a two dimensional grid by means of four possible
actions, left, down, right, and up.
There many variations, e.g.
\cite{boyan1995generalization,crook2003learning, melnikov2014projective}.

In this paper all Gridworld MDPs will
be of size $7*7$ with walls surrounding the space.  
Any attempt to move into a wall location will instead leave
the agent in its original location.  There is one reward location, and a game episode
ends when the reward is collected or after 100 moves.  Actions are deterministic as this makes
generalization more difficult.  (It has been
repeated observed (e.g., \cite{cobbe2018quantifying,mehta2019active})
that adding randomization makes extrapolation easier, since the more  states
observed at training time the fewer unobserved states can occur later.)
The start and goal states may vary, but must be distinct.
To keep our Gridworld particularly simple there are no interior walls..
See
\begin{figure}
\begin{tabular}{c|c|c|c|c|c|c|c|} 
   & 0  & 1 & 2 & 3 &  4 & 5 & 6  \\ \hline
0 & X & X & X & X & X & X & X  \\ \hline
1 & X & * &     &      &    &    & X \\ \hline
2 & X &   &   &   &    &   & X \\ \hline 
3 & X &   &   &   &    &   & X \\ \hline 
4 & X &   &   &   & @ &   & X \\ \hline 
5 & X &   &   &   &   &   & X \\ \hline 
6 & X & X & X & X & X & X & X  \\ \hline
\end{tabular}
\caption{Gridworld game, where ``@'' represents the agent, which receives a reward if it gets
to the position marked with ``*''. }
\label{gridworld}
\end{figure}
Figure \ref{gridworld}

Instances of Gridworld are parameterized by the start and reward locations.
We create some number of training instances (1, 2, 4, 8, or 16)  and 10
testing instances. 
All test instances must
have as their reward state a location not used for the reward in  any the training cases, thus ensuring all states encountered
in test trajectories, including the start state, will be unreachable.  Also we ensure there are no duplicate game instances within  the
training set or testing set.
Since values of
our various metrics are functions of the randomized choices of training and testing instances, each
observation is made 20 times and we report the mean.  

The function-approximation method for computing Q-values, or policy-action-probability logits, is shown in
\begin{figure}
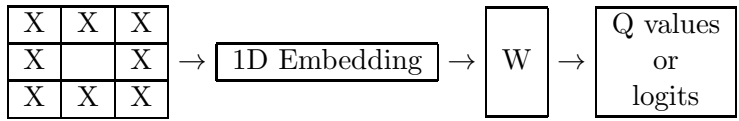

\begin{tabular}{|c|c|c|} \hline
X & X & X \\ \hline
X & & X \\ \hline
X & X & X \\ \hline
\end{tabular} $\rightarrow$
\begin{tabular}{|c|} \hline
1D Embedding \\ \hline
\end{tabular} $\rightarrow$
\begin{tabular}{|c|} \hline
 \\
W \\
\\ \hline
\end{tabular} $\rightarrow$
\begin{tabular}{|c|} \hline
Q values \\
or \\
logits \\ \hline
\end{tabular}
\caption{Basic function approximation of Q values or policy-probability logits\label{nonlin}}
\end{figure}
Figure \ref{nonlin}.The input is a $7*7$ array of indices for the four kinds of objects that
can populate a location: space, agent, star, and wall.  (When we wish to be more general we let $g$ be the one dimensional size of
the grid, so $g=7$.)  The object indices are immediately converted
to trainable embeddings.  Then  a single 
full width matrix W converts the input to Q values  (for Q function learning methods) or,  for policy-gradient-learning logits  that are feed into softmax.
The loss function for Q-value-function learning is squared distance between
the current value estimate and the improved estimate from computing the next reward  plus
the maximum estimated Q value for the next state.
In all cases we use experience replay.\cite{schaul2015prioritized}  

For hyperparameter details, see Appendix A

\section{Deep Q Learning vs. REINFORCE\label{dqnrei}}

\begin{table}
\begin{tabular}{c|c|c|c}
\# train  & \% comp. train & \# over min. train & \% comp. test \\
16 & 0.72 & 11.3  &  0.05  \\
 8 & .78 & 9.0 & 0.07 \\
 4 & 0.93 & 5.5 & 0.09 \\
 2 &  0.95 & 2.66 & 0.05 \\
1 & 1.0 & 0.93 & 0.07
\end{tabular}
\caption{\label{dqn}Performance of deep Q learning on Gridword with differing  number of training instances}
\end{table}
Table \ref{dqn}
shows how well our deep Q learning version of Gridword is able to generalize,  A constant of generalization research is that the more variation 
we see at test time the better the generalization. For example, \cite{cobbe2018quantifying} invent a new Atari-style video game, ``Coin Run,''  with which they can create any
arbitrary number of new versions, all of which have unreachable states.  As the number of train-time game  instances grows from 100 to 10,000 they
observe that the percentage of new test-time  games that are successfully completed grows.  At the same  time, the percentage of train time solutions
goes down. That is, as the training regime gets more difficulty, the program does less well at training time, but more caries over to test time.

We see some of this in Figure \ref{dqn} except first, rather than trying $10^2$ to $10^4$ test instances  we leverage the simplicity of our Gridworld
to reduce this to 16 down to 1, as the label in the first column in each row indicates.  The second  and third columns have two measures of train-time performance.  Column two indicates fraction of train-time completion, which increases monotonically from 77\% with 16 train-time instances, to 100\%
with one or two such instances.  Column three shows for the games that are completed, how many steps the completion took, on average, over the
minimum (mistake-free) trajectory length.  This too shows improving train-time performance as the variation seen at train-time decreases.

However, the test-time performance illustrated in Table \ref{dqn} is quite different.  To a first approximation  {\em none} of the test-time games
are solved.  A look at what happens in these games quickly illustrates that in virtually all cases the policy created has an infinite loop for one
or more steps in  all of 200 testing instances.  (We test on 10 game instances, all of which start in a unreachable state.  We run the entire train-test
cycle 20 times, creating a new suite of training and testing games, and report mean results).  E. g. in position (3,4) the maximum Q-value corresponds to moving up, but in position(2,4) it is for
moving down.  The key point here is that standard Q function learning  at test time has us always deterministically choosing the move with maximum Q-value in the current state,
so any inaccuracy  in the Q-value/state association is blown up by the determinism.\footnote{We have done experiments that combine Q-learning
with a temperature guided probabilistic choice of actions at test-time. While this improves extrapolation greatly, we have observed odd behavior using
this scheme that we have not been able to diagnose, and thus have put it aside in this study.}

Secondly there are lots of reasons to expect inaccurate associations.  The input  state representation is simply our 2D grid so at test time, if
our agent is in position (3,4), the most similar input states are those with the agent at this position {\em no matter where the reward state is
located} since in all such positions the difference is in one grid location.  So to a first approximation the policy at (3,4) is going to be independent of
the testing reward location.  In particular, when trained on a single goal, the agent will still head toward the single training goal, not
the testing one.  When trained on 16 various  goals, it will head
toward whichever of the goals happens to be ascendant, presumably by some small amount.  No doubt with a much-larger capacity network, and a lot
more training, a Q-learner could recognize the significance of reward-state position.  But the goals of this paper is to make unreachable state generalization
easy, not to beat it with a sledge hammer.

\begin{table}
\begin{tabular}{c|c|c|c}
\# train & \% comp test & \# over min. test & \% trivially wrong \\
16 & 0.28 & 14.55 & 0.12  \\
8 & 0.24 & 13.25 & 0.13 \\
4 & 0.26 & 16.51 & 0.14  \\
2 & 0.2 & 15.54 & 0.17  \\
1 & 0.18 & 8.89 & 0.18 
\end{tabular}
\caption{\label{reinforce}Extrapolation with REINFORCE.  \% completed, \# moves over minimum, percentage trivially wrong} 
\end{table}
Table \ref{reinforce} shows Gridworld extrapolation when using REINFORCE\cite{williams1992simple}.  Again we consider training on 16 down to
1 game instance.  While extrapolation ability is poor, at its best 28\% of test games complete, it is not zero.  

However, the policies still
have not learned much about Gridworld.  The final column  of Table \ref{reinforce} is labeled ``trivially wrong'' and shows the percentage of
probability  mass, averaged over all states in all game instances, devoted to moves that can be determined to be sub-optimal simply by looking
at the four board positions surrounding the agent.

For example, suppose the agent is in position (3,1), of Figure \ref{gridworld} (the middle left-hand side of the grid, one to the right of the left-hand wall)
and the goal state
is at (4,4) or any other place not directly next to the agent.  We say that a move left (into the wall) is trivially wrong, but the other three moves
are not.  If the test-time policy we have learned assigns 18\% probability to moving left, then the trivially wrong score for this state/policy
combination is 18\%.  If there were two  adjacent walls, then the trivially-wrong score would be the sum of the probabilities assigned to the
two moves.  If there were an adjacent goal, then all three non-goal moves would be trivially wrong.

For initial policies, where  all moves have probability  25\% plus or minus some jitter, the trivially wrong score will be about 22\% depending slightly
on the average positions of the goal states.  So Table \ref{reinforce} shows with 4 or more game instances for training we can lower the trivially-wrong rate by about half.  It has learned
something, but not much.  Nevertheless, it seems clear that policy gradient methods are superior to Q-learning as a basis for generalization research and
REINFORCE is used in all following experiments.

\section{Ego-Centric Grid Representation\label{egosec}}

An {\em ego-centric grid representation}, also called {\em agent-centric} or {\em deictic} representation (\cite{agre1987pengi,ravindran2004algebraic,
konidaris2012transfer,finney2012thing,james2019learning}) is one in which the representation
depends on the agent's point of view.  In this paper we adopt a very simple version of this
in which rather that placing the origin of the Gridworld at an arbitrary point, e.g., the upper left-hand corner, we instead make the origin always the
position of the agent.  Or as we will show it, the agent is always at the center of the grid.  So
the grid of Figure \ref{gridworld} would now look like 
\begin{figure}
\begin{tabular}{c|c|c|c|c|c|c|c|c|c|c|c|}
& 0 & 1 & 2 & 3 & 4 & 5 & 6 & 7 & 8 & 9 & 10  \\ \hline
0 &  &   &   &   &   &   &   &   &   &   &   \\ \hline 
1 &   & X & X & X & X & X & X & X &   &   &   \\ \hline 
2 &  & X & *  &    &     &    &    & X  &   &   &   \\ \hline 
3 &  & X &    &    &     &    &    & X  &   &   &    \\ \hline 
4 &   & X &   &   &   &     &   & X &   &   &    \\ \hline 
5 &   & X &   &   &   & @ &   & X &   &   &    \\ \hline 
6 &   & X &   &   &   &   &   & X &   &   &   \\ \hline 
7 &   & X & X & X & X & X & X & X &   &   &   \\ \hline
8 &   &   &   &   &   &   &   &   &   &   &    \\ \hline
9 &   &   &   &   &   &   &   &   &   &   &    \\ \hline
10 &  &   &   &   &   &   &   &   &   &   &   \\ \hline 
\end{tabular}
\caption{\label{egocentric} The Gridworld of Figure \ref{gridworld} expressed in ego-centric coordinates}
\end{figure}
Figure \ref{egocentric}.

Notice how the agent is at location (5, 5), and no matter where it moves in the smaller 7*7 space, it remains at location (5,5) in ego-centric space.
We have also padded both the X and Y
axes with 4 extra spaces.  This allows the representation to put the agent in the center and still 
show the complete original grid, even if in the original grid the agent was in one of the extreme corners.
(We let $x$ be the one dimensional size of the ego-centric array, so given that the agent is barred from rows and columns 0 and 6, it can
be found that to make sure every bit of the original is visible in the ego-centric version $x = 2g-3$.)

At first this looks like a bad move, since ego-centrism nearly doubles the linear dimension of our grid, and
thus increases the number of weights in the matrix that turns the grid representation into actions by
nearly a factor of 4
(see Figure \ref{nonlin}). However, we now argue that it is well worth the cost

\begin{figure}
\begin{tabular}{c|c|c|c|c|c|c|c|} 
   & 0 & 1 & 2 & 3 & 4 & 5 & 6 \\ \hline
0 & X & X & X & X & X & X & X \\ \hline
1 & X &   &   &   &   &   & X \\ \hline
2 & X & @ &   &   &   &   & X \\ \hline
3 & X &   &   &   &   &   & X \\ \hline
4 & X & @ &   &   &   &   & X \\ \hline
5 & X &   &   &   &   &   & X \\ \hline
6 & X & X & X & X & X & X & X \\ \hline 
\end{tabular}
\caption{\label{twosit} Superimposed Gridwords showing where agent appears in two cases}
\end{figure}
Figure \ref{twosit} superimposes two game states which are similar insofar as in both cases the
agent should not move left (move 0) because of the wall in that location.  Consider training
on the first case, (don't move into wall position from (2,1) but concentrate on  what the agent picks up about what it should do in the second (don't move
left from (4,1)). As shown
in Figure \ref{nonlin} we turn our board into a vector of size 49, and then multiply it times
a $49*4$ weight matrix to turn the board into logits for the 4 possible moves.  (Actually,
we also have an embedding size of 2, which double these numbers, but to keep things simple, let's
assume an embedding size of 1.)  The 1D board vector now looks like this (again with both
agent positions superimposed on the same vector.
\begin{quote}
\begin{tabular}{|c|c|c|c|c|c|c|c|c|c|c|}
0 & 1 & 2 & $\cdots$ & 14 & 15 & $\cdots$ & 28 & 29 & $\cdots$ & 48  \\ \hline
X & X & X &               & X  & @  &         & X  & @  &          & X \\ \hline
\end{tabular}
\end{quote}
When we learn not to go into the wall from (2,1) the agent position  corresponds to position 15 and the change in $W$ will  be specific to this location,
There will be little to nothing that translates into information about what to do when the agent is at location (4,1,) (or position 29 in the vector board representation).

Contrast this with the processing when we use an ego-centric representation.  In both cases we see
the following piece of the state representation:
\begin{quote}
\begin{tabular}{|c|c|c|c|c|c|c|}ß
0 & 1 & $\cdots$ & 59   &  60   & $\cdots$ & 120  \\ \hline
  &     &               & X & @ &            &   \\ \hline  
\end{tabular}
\end{quote}
So modifying the weights in case one automatically help in case two. Or to put it more succinctly,
ego-centric representation ought to help in RL extrapolation. And in Table \ref{egores} we see that
indeed it does.\footnote{It might be noted/objected that something like the same generalization
could be obtained by using convolutional filters on the original grid since convolution automatically
induces translation invariance  But it is not to hard to convince yourself that translation
invariance  is not really right for Gridworld problems.  We see a vertical  wall in positions
(0, 0:6) and (6, 0:6) but not elsewhere.  Similarly if we see the agent in one location, it is excluded from
all others at that time instant.}
\begin{table}
\begin{tabular}{c|c|c|c}
\# train & \% comp test & \# over min. test & \% trivially wrong \\
16 & 0.9 & 2.29 & 0.03  \\
8 & 0.93 & 3.23 & 0.03 \\
4 & 0.88 & 5.23 & 0.05 \\
2 & 0.62 & 6.31 & 0.1\\
1 & 0.35 & 7.98 & 0.16
\end{tabular}
\caption{Extrapolation results with ego-centric coordinates\label{egores}}
\end{table}

Table \ref{egores}  shows our extrapolation results when using ego-centric coordinates.  Contrast them with those in Table \ref{reinforce}.  Most
notably the completion percentages have more than doubled (32\% down to 10\% without ego-centrism, 84\% down to 24\% with).  The quality of
the solutions are up (see number over minimum number of moves necessary) and from the probability mass devoted to trivially wrong moves
(e.g., 3.6\% when trained on 16 instances) it is clear that 
the policy has learned to handle the basics reasonably well, when trained on 16, 8, or 4 instances.

\section{Symmetry\label{symsec}}

In the last section we saw how ego-centric spatial representation allows an agent to automatically
generalize between some kinds of learning experiences --- those which look similar when viewed from
the agent's point of view.  There are other cases, however that ego-centrism does not catch --- for example,
learning not to move right when there is a wall to your right does not help to learn not to move left
when there is a wall to the left.  Or more generally, Gridworld is inherently  90 degrees  rotationally symmetric but nothing we have
done so far captures this.

Exploiting symmetry has been occasionally used in computer vision work, but requests for papers about
computer vision and symmetry almost exclusively return papers on recognizing objects with symmetry.  One
exception is \cite{dieleman2015rotation}, who applied convolutional deep learning to the classification
of galaxies.  Pictures of galaxies do, of course, have rotational symmetry, and this property is added
in \cite{dieleman2015rotation} by data augmentation. The galactic photographs are repeated in the data
set with different 90 degree rotations  Also rotation  is mentioned in \cite{cobbe2018quantifying} but in the immediate context of image augmentation,
and no study of its use in RL is presented.   
\cite{marcos2016learning} directly build rotational convolution  into their textture recognition system as texture is another area where images
show rotational symmetry.   However, they do this by enforcing rotational symmetry on their convolutional kernels, which makes sense for texture, but 
does not work for us. A theoretical analysis  of symmetry in RL  can be found in \cite{ravindran2004algebraic}.

Here we too build rotational (and later mirror image) symmetry directly into our learning mechanism, but in a novel fashion.
\begin{figure}
\begin{tabular}{|c|c|c|c|c|c|c|c|c|c|c|} \hline
  &   &   &   &   &   &   &   &   &   & o  \\ \hline
  &   &   &   &   &   &   &   &   & o & o   \\ \hline
  &   &   &   &   &   &   &   & o & o & o  \\ \hline
  &   &   &   &   &   &   & X & o & o & o  \\ \hline
  &   &   &   &   &   & o & X & o & o & o \\ \hline
  &   &   &   &   & @ & o & X & o & o & o  \\ \hline
  &   &   &   &   &   & o & X & o & o & o  \\ \hline
  &   &   &   &   &   &   & X & o & o & o  \\ \hline
  &   &   &   &   &   &   &   & o & o & o  \\ \hline
  &   &   &   &   &   &   &   &   & o & o  \\ \hline
  &   &   &   &   &   &   &   &   &   & o  \\ \hline
\end{tabular}
\caption{Triangular-quadrant two, $t(2)$, array values for Figure \ref{egocentric}\label{triquad}. The triangular quadrants are labeled  0-3, in correspondence with the directions of motion 0-3. Here blank space
indicates grid positions not in $t(2)$, and ``o'' indicates inclusion  in $t(2)$ and there is a space at the location.}
\end{figure}
Standard convolution builds translational symmetry into visual processing by translating patches to line up with a kernel 
before taking the dot product.  We do the same, but rather than a translation/dot-product process we use a rotation/dot-product one.  The shape of
the kernel is shown in 
Figure \ref{triquad}.  To get full coverage of the image we rotate it four times to bring each quarter into alignment with our pattern. 
The idea is that the matrix $W$ of Figure \ref{nonlin} should not cover the entire $x*x$ Gridworld matrix but rather just a quarter of it.\footnote{
To be more precise, since a symmetric form for our quadrant requires
overlap between the triangular shapes (at the origin, and along one of the diagonals) four copies of the kernel shape cover $x^2$ spaces, plus
$4*(x-1)/2$ (4 extra diagonals) + 1 (and extra extra copy of the origin) = $x^2 + 2x -3$ parameters.}

We capture mirror symmetry in a similar fashion.  Now the kernal only examines, say, the bottom half of the quadrant in Figure \ref{triquad}, as
shown in 
\begin{figure}
\begin{tabular}{|c|c|c|c|c|c|c|c|c|c|c|c} \hline¯
  &   &   &   &   &   &   &   &   &   &   \\ \hline
  &   &   &   &   &   &   &   &   &  &    \\ \hline
  &   &   &   &   &   &   &   &   &   &    \\ \hline
  &   &   &   &   &   &   &   &   &   &    \\ \hline
  &   &   &   &   &   &   &   &   &   &   \\ \hline
  &   &   &   &   & @ & o & X & o & o & o  \\ \hline
  &   &   &   &   &   & o & X & o & o & o  \\ \hline
  &   &   &   &   &   &   & X & o & o & o  \\ \hline
  &   &   &   &   &   &   &   & o & o & o  \\ \hline
  &   &   &   &   &   &   &   &   & o & o  \\ \hline
  &   &   &   &   &   &   &   &   &   & o  \\ \hline
\end{tabular}
\caption{A triangular octant for Figure \ref{egocentric}\label{trioct}}
\end{figure}
Figure \ref{trioct} and we have to perform both rotations and inversions to bring all pieces of our image into alignment.

\begin{table}
\begin{tabular}{c|c|c|c|c}
\# train & \% comp test & \# over min. test & \% trivially wrong & imbalance \\
16 & 1.0 & 0.14 & 0.0 & 0.81 \\
8 & 1.0 & 0.23 & 0.0 & 0.8 \\
4 & 0.97 & 1.06 & 0.02 & 0.76 \\
2 & 0.91 & 1.05 & 0.03 & 0.73 \\
1 & 0.89 & 1.49 & 0.02 & 0.7 
\end{tabular}
\caption{\label{symres} Extrapolation results with built-in symmetry}
\end{table}
The extrapolation results obtained by building rotational and mirror symmetry are shown in Table \ref{symres}. They show a marked improvement
over those for ego-centric co-ordinates alone.  Completion is now high, even when trained on a single example, and the trivially-wrong rates
are low in all cases.

\section{Maximizing Action-Choice  Entropy\label{entsec}}

Policy-gradient RL methods in general, and REINFORCE in particular do not necessarily converge to a unique optimum when two or more actions
both engender   the maximum discounted reward.  In such cases any combination of the best moves summing to one produce an
equally good policy ---  for the training examples.   However from a generalization point of view they are not equally good.  
The more states an agent
encounters in training the fewer new ones it will stumble across in testing, so everything else equal, we prefer policies with maximum action entropy.

Unfortunately this is not the outcome we standardly get as  a result from REINFORCE, or, as far as we can tell, from any of the other policy
gradient methods.  Looking again at Table \ref{symres} we see a new column labeled ``imbalance''.  For every position
in our game trajectory for which two moves are equally good we record the absolute value of the difference between the probabilities our policy
assigns to the two moves and then average over all of these situations.  (If there is only one best move we record nothing.)  The average
is listed in the ``imbalance'' column.  Note if in all situations the two optimum moves had equal policy probability the number would be zero.
If they all assigned all of their probability mass to one of the two, then the number would be one.  If the chance of any divide were equally likely we would
expect to see an average imbalance of .5.  
We observe none of these outcomes, but imbalances  are very much higher than one would expect from a random distribution of outcomes.  That is, they
are far from the maximum entropy distribution over  actions that we would prefer everything else equal.  We assume this is related
to the failure of the iid-assumption.

Many researchers have added
a second, entropy, term to the loss in order to increase the training time state coverage, and we do so here.  The results are shown
\begin{table}
\begin{tabular}{c|c|c|c|c}
\# train & \% comp test & \# over min. test & \% trivially wrong & imbalance \\
16 & 1.0 & 0.28 & 0.02 & 0.65 \\
8 & 1.0 & 0.58 & 0.01 & 0.67 \\
4 & 1.0 & 0.57 & 0.02 & 0.61 \\
2 & 0.96 & 1.62 & 0.02 & 0.59 \\
1 & 0.94 & 3.39 & 0.07 & 0.54 
\end{tabular}
\caption{\label{entres}Extrapolation results with secondary minimum entropy loss}
\end{table}
Figure \ref{entres}.  We also made the mixing factor between the two desiderata a decreasing function of episode number as without that we found
a pronounced negative effect on the number of moves over minimum (it went up), and the percentage of trivially wrong moves (again it went up).
Furthermore,  as we can see from the last column in Figure \ref{entres},  the imbalance is mitigated to only a small degree.\footnote{One
might hope that actor-critic methods such as a2c\cite{mnih2016asynchronous} would fix this problem since they explicitly compute the value
function, and, after all, the value of the states resulting from equally good moves are equal.  However, our experiments indicate this is not the
case.  While the values for the moves end up more or less equal, this does not change the dynamics of the policy probability update mechanism
any more than do the correct discounted rewards when using REINFORCE.}

While Table  \ref{entres} does show improvements in completion rate, we include this section more to emphasize what we consider
the negative interactions between policy-gradient methods and state coverage an unsolved problem.

\section{Conclusion}
We have presented results  on extrapolation in Gridworld MDPs.  In particular we conclude that
\begin{itemize}
\item policy gradient methods extrapolate better than ones using estimates of Q values.  First, as seen in many other contexts, it seems to be easier to learn a policy directly than to estimate Q-values, and more specifically, the Q-value optimization step of deterministically taking the move that leads to maximum
state value makes extrapolation very difficult.  
\item ego-centric space representation makes differences since similarities in spatial configurations better align to those that need attention for move choice  --- e.g., a wall to the left looks nearly identical for various agent positions, thus bringing into play the same deep-learning parameters.
\item building  rotational and mirror-image symmetry into the learning mechanism helps a great deal as it goes much further in unifying  the response to common situations (the same parameters deal with moving left to the goal state as with moving down).  Furthermore this can be accomplished in
a relatively straight-forward fashion using rotational and mirror convolution rather than standard translational convolution.
\item all of policy-gradient methods we have tried have a very strong tendency to find nearly minimum-entropy action choices when two or more
actions are equally good, leading to fewer well explored states at training time.  Furthermore we find that adding a minimum entropy loss has only a
very small effect on completion rate, while noticeably worsening quality of solutions.  
\end{itemize}
To boil this down even further, after we moved from Q-value learning to policy-gradient methods, the most important modifications we found
were ones to the representation of the current state.  Once ``knowledge representation'' was a standard sub-area of artificial intelligence.  Perhaps
reinforcement learning could benefit from its resuscitation. 

\bibliography{rlbib}{}
\bibliographystyle{alpha}
\appendix
\section{Hyperparameters}
We use an embedding size of 2 for our 4 objects: space, wall, agent, goal).  To speed convergence in training embeddings 
are initialized to (0,0) for space, and
three of the four corners of the space (+/- .1, 0), (0, +/- .1).  For game reuse we maintain a set of 228 full games.  At every epoch the
oldest 32 are removed and a new 32, based upon the current policy, are added. Games are terminated when the goal state is achieved
or after 100 moves.  

A training epoch consists of 3*NumMoves random selections from the set of moves. There are 200 epochs.  We used the Adam optimizer with a learning
rate of .002, and a batch size of 10.  Gradients were clipped at +/-20.

We simplified the maximum entropy calculations in Section 6 by adding a fraction $f$ of the highest probability to the loss.  $f$ is initially .25, and
it is reduced by .99 at each epoch.

\end{document}